\pdfoutput=1

\documentclass[11pt]{article}

\usepackage[preprint]{acl}

\usepackage{times}
\usepackage{latexsym}

\usepackage[T1]{fontenc}

\usepackage[utf8]{inputenc}

\usepackage{microtype}

\usepackage{inconsolata}

\usepackage{graphicx}
\usepackage{booktabs}
\usepackage{subcaption}

\title{MeDiSumQA: Patient-Oriented Question-Answer Generation from Discharge Letters}

\author{
 \textbf{Amin Dada\textsuperscript{1}},
 \textbf{Osman Alperen Koraş\textsuperscript{1}},
 \textbf{Marie Bauer\textsuperscript{1}},
 \textbf{Amanda Butler Contreras\textsuperscript{2}},
\\
 \textbf{Kaleb E Smith\textsuperscript{2}},
 \textbf{Jens Kleesiek\textsuperscript{1,3,4,5}},
 \textbf{Julian Friedrich\textsuperscript{1}},
\\
 \textsuperscript{1}Institute for AI in Medicine (IKIM), University Hospital Essen, Germany\\
 \textsuperscript{2}NVIDIA, Santa Clara, USA\\
 \textsuperscript{3}Cancer Research Center Cologne Essen (CCCE), University Medicine Essen, Germany\\
 \textsuperscript{4}German Cancer Consortium (DKTK, Partner site Essen), Germany\\  
 \textsuperscript{5}Department of Physics, TU Dortmund, Germany
\\
 \small{
   \textbf{Correspondence:} \href{mailto:amin.dada@uk-essen.de}{amin.dada@uk-essen.de}
 }
}

\begin{document}
\maketitle
\begin{abstract}
While increasing patients' access to medical documents improves medical care, this benefit is limited by varying health literacy levels and complex medical terminology. Large language models (LLMs) offer solutions by simplifying medical information. However, evaluating LLMs for safe and patient-friendly text generation is difficult due to the lack of standardized evaluation resources. To fill this gap, we developed \textbf{MeDiSumQA}. \textbf{MeDiSumQA} is a dataset created from MIMIC-IV discharge summaries through an automated pipeline combining LLM-based question-answer generation with manual quality checks. We use this dataset to evaluate various LLMs on patient-oriented question-answering. Our findings reveal that general-purpose LLMs frequently surpass biomedical-adapted models, while automated metrics correlate with human judgment. By releasing \textbf{MeDiSumQA} on PhysioNet, we aim to advance the development of LLMs to enhance patient understanding and ultimately improve care outcomes.
\end{abstract}

\section{Introduction}
Access to health documents empowers patients and improves medical care \citep{greene2012does, lye201821st, ross2003effects}. These documents, however, often use language too complex for patients to understand \citep{paasche2005prevalence}, and physicians have no time to simplify documents in a patient-friendly manner \citep{ammenwerth2009time}.

This gap between healthcare providers and patients can be bridged by large language models (LLMs) \citep{ali2023using, jeblick2024chatgpt, zaretsky2024generative, eisinger2025s}. Through their ability to simplify medical information, LLMs can enhance the access to health documents and ultimately improve patient care.
However, assessing and comparing LLMs in their ability to generate safe and patient-friendly text remains challenging due to the lack of benchmarks and publicly available resources. Strict privacy regulations surrounding clinical data limit dataset accessibility, thereby impeding the development of open benchmarks for evaluating LLMs in medical contexts.

 \begin{figure*}[ht]
  \centering
  \includegraphics[width=\textwidth]{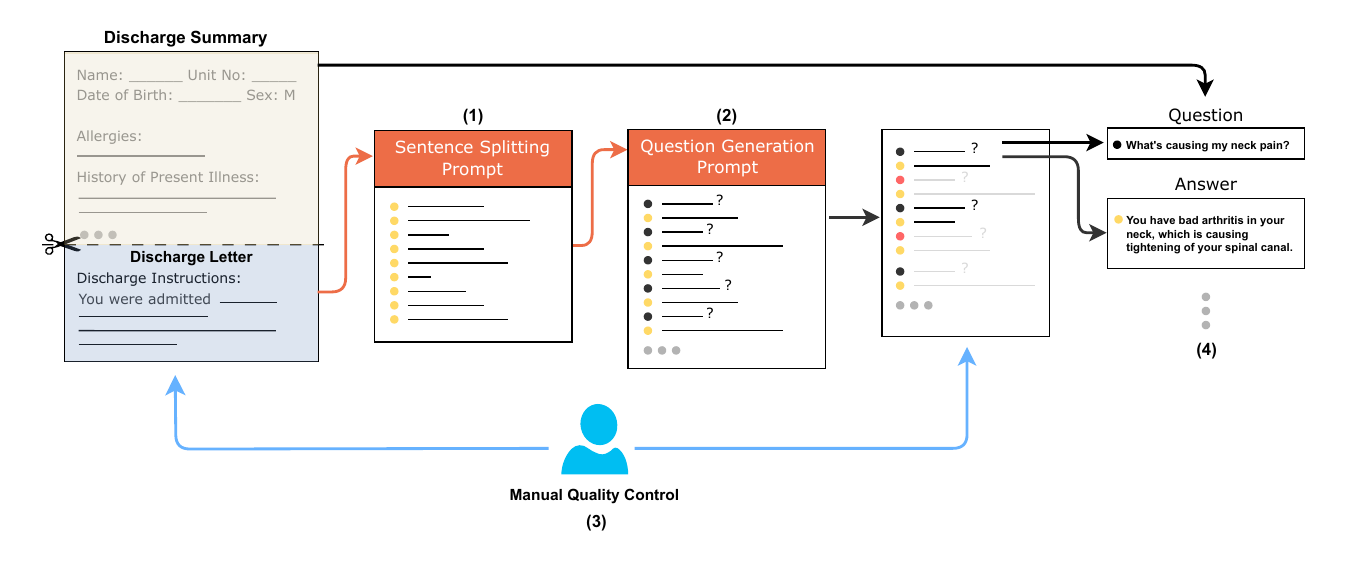}
  \caption{Generation process of \textbf{MeDiSumQA}. After identifying the discharge letter, we separate it from the main document and use an LLM to split it into sentences (1). Based on these sentences, we let an LLM generate matching questions (2). The resulting question-answer pairs were reviewed and curated by a physician, resulting in the the final \textbf{MeDiSumQA} dataset of 416 question-answer pairs (3). For inference, we provide LLMs with the discharge summary (without the bottom discharge letter) and pose the generated question. The model answer is then compared to the extracted ground truth answer (4).}
  \label{medisumqa_gen}
\end{figure*}

 To address this issue, we developed \textbf{MeDiSumQA}. \textbf{MeDiSumQA} is a novel, patient-oriented question-answering (QA) dataset, a format especially suitable to improve patient understanding of clinical documents \citep{cai2023paniniqa}. 

In this paper, we describe how we created, curated, and evaluated \textbf{MeDiSumQA}, crafting a standardized resource for future benchmarking. By making this task openly available to researchers, we support broader development and testing of LLMs for healthcare applications, helping address challenges of time constraints and health literacy. 

\section{Related Work}

While several clinical QA datasets exist \citep{pampari-etal-2018-emrqa, lehman-etal-2022-learning, soni-etal-2022-radqa, bardhan-etal-2022-drugehrqa, dada2024information, kweon2024ehrnoteqa}, none, to the best of our knowledge, are explicitly designed for patient-oriented use. 

Prior research has explored medical text simplification, but did not focus on helping patients understand clinical documents in a QA format. \citet{aali2024dataset} developed a public dataset that converts MIMIC hospital course summaries into concise discharge letters. \citet{campillos2022building} created a Spanish dataset for simplifying clinical trial texts, demonstrating the importance of multilingual resources. \citet{trienes-etal-2022-patient} focused on making pathology reports more understandable for patients, though their dataset remains private and does not address everyday clinical questions. Similarly, while \citet{ben-abacha-demner-fushman-2019-summarization}'s MeQSum dataset transforms consumer health questions into brief medical queries, it lacks strong clinical focus.

Our work addresses these limitations by introducing a public, patient-centered QA dataset based on clinical MIMIC-IV discharge summaries, creating a benchmark to evaluate LLMs.

\section{Methods}

\subsection{Dataset Generation}

In the MIMIC-IV dataset \citep{johnson2023mimic}, some discharge summaries conclude with a discharge letter that summarizes key information and follow-up instructions in patient-friendly language. We used these discharge letters as the foundation for generating QA pairs in the following manner (Figure \ref{medisumqa_gen}):

First, we identified discharge summaries containing discharge letters by searching for the string\footnote{``You were admitted to the hospital''} that indicates the start of a discharge letter.
We split each discharge letter into sentences using \textit{Meta's Llama-3-70B-Instruct} \cite{dubey2024llama}, which proved more accurate than traditional sentence splitters like NLTK, especially when handling irregular formatting and placeholders introduced by anonymization. To ensure accuracy, we prompted the LLM to preserve the original sentence structure and wording, which we subsequently verified by confirming that each processed sentence could be matched exactly with its source in the original discharge letter via exact string matching.
 
Afterwards, we fed these sentences into an LLM to generate matching questions from a patient's perspective. The LLM was allowed to reformulate the answer to match the question, but was instructed to reference the source sentence. We then checked these references to confirm that no information from the source document was altered. Since the answers are directly derived from the discharge letters written by medical professionals, this method maintains both medical accuracy and patient-friendly language. All mentioned prompts are listed in Appendix \ref{sec:appendix:prompts}.

 The resulting QA candidates were then manually reviewed by a physician who selected high-quality examples based on the following criteria:

    \begin{figure*}[ht]
  \centering
  \includegraphics[width=0.9\textwidth]{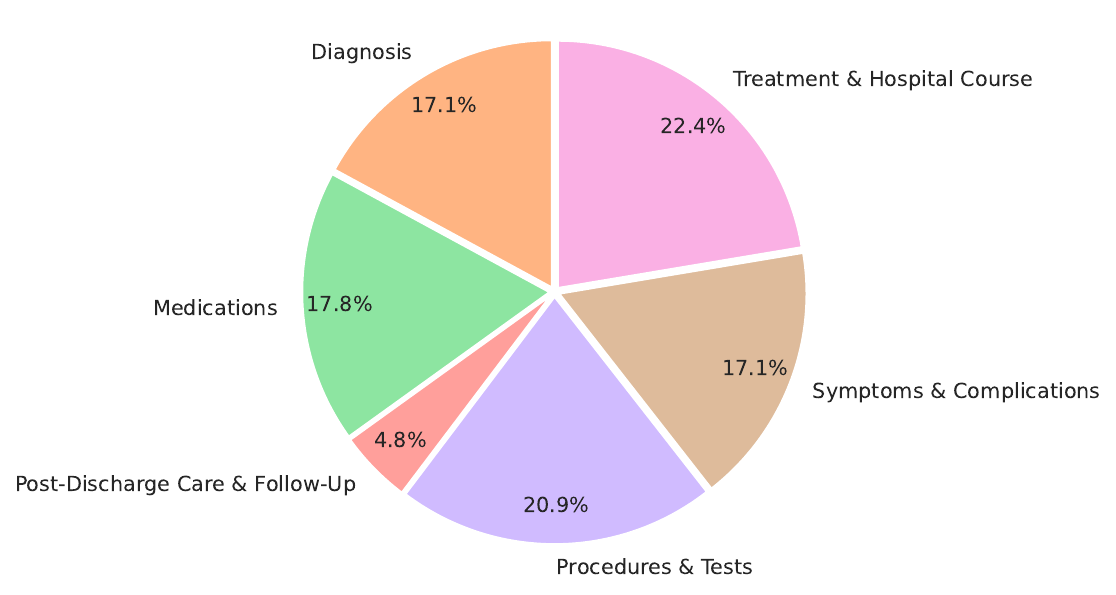}
  \caption{Frequency of question-answer categories in  \textbf{MeDiSumQA}.}
  \label{medisumqa_categories}
\end{figure*}
\begin{enumerate}
    \item[] \textbf{Factual correctness} Question-answer pairs had to be logically connected. Answers that did not match their questions (e.g., "What medication should I avoid taking due to a possible allergy?" - "You were prescribed ibuprofen") were excluded.
    \item[] \textbf{Completeness} Answers had to be complete. Partial answers (e.g., "What medications were started for me?" - "You were started on Vancomycin 1gm IV every 24 hours" when additional antibiotics were prescribed) were discarded.
    \item[] \textbf{Safety} Answers needed to be safe. Potentially harmful instructions (e.g., "Take Coumadin 3 mg daily" without mentioning INR monitoring) were excluded.
    \item[] \textbf{Consistency} Questions had to be answerable from both the discharge letter and discharge summary. Questions whose answers relied solely on information from the discharge letter were excluded.
    \item[] \textbf{Complexity} Question-answer pairs had to be sufficiently complex. Obvious answers or overly specific questions that gave the answer away (e.g. "Did I receive Ciprofloxacin?" - "You received Ciprofloxacin.") were excluded.
\end{enumerate}

As a final step, we removed the discharge letters from their summaries and combined the remaining summaries with their matching QA pairs. This resulted in three components, forming \textbf{MeDiSumQA}:

\begin{enumerate}
    \item \textbf{A question} that serves as input for LLMs.
    \item \textbf{An abbreviated discharge} summary without the discharge letter that LLMs use to answer the input question
    \item \textbf{A ground truth answer} for comparison with generated responses
\end{enumerate}

\subsection{QA Categories}

In \textbf{MeDiSumQA}, we identified six QA categories:
\begin{itemize}
    \item Symptoms \& Complications
    \item Procedures \& Tests
    \item Diagnosis
    \item Treatment \& Hospital Course
    \item Medications
    \item Post-Discharge Care \& Follow-Up
\end{itemize} 
To assign each QA pair to one of these categories, we used Meta's \textit{Llama-3.3-70B-Instruct} \citep{dubey2024llama}.

\subsection{Evaluation}

\begin{figure*}[h]
  \centering
  \includegraphics[width=0.8\textwidth]{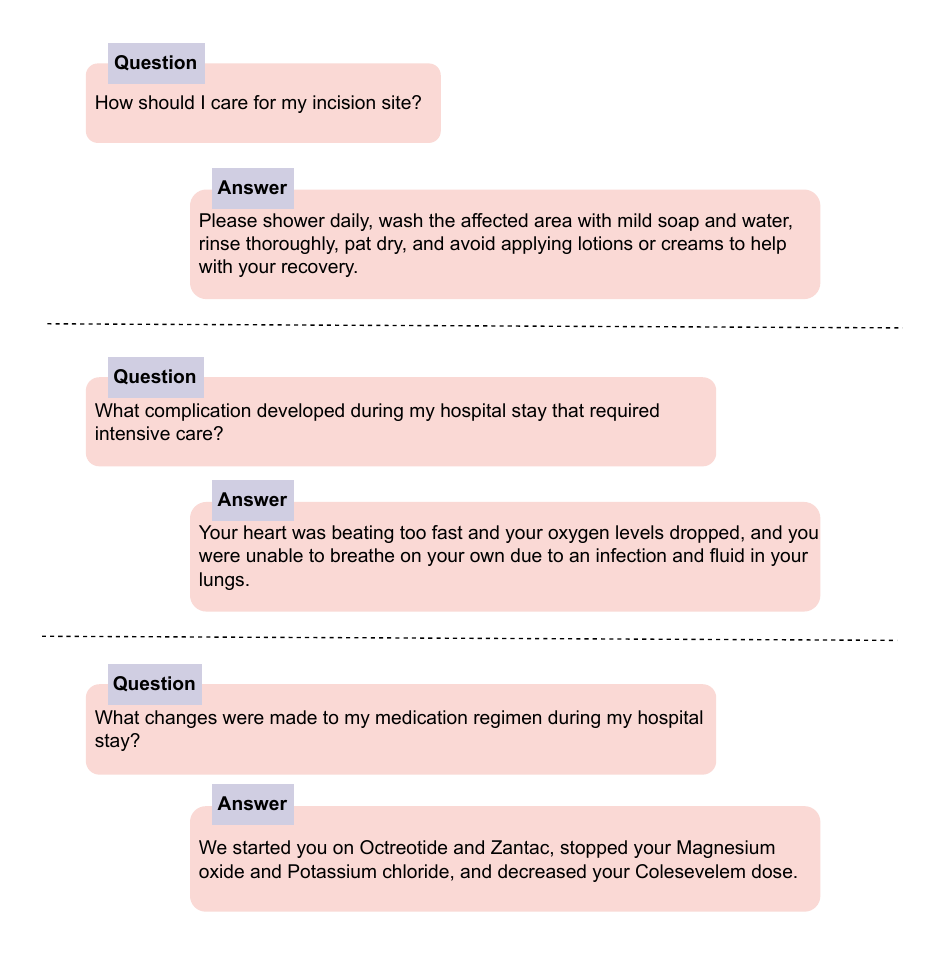}
  \caption{Example of QA pairs in \textbf{MeDiSumQA} dataset.}
  \label{qas_examples}
\end{figure*}

We evaluated the following models on \textbf{MeDiSumQA}: \textit{Mistral-7B-Instruct-v0.1} \citep{jiang2023mistral}, \textit{Meta-Llama-3-8B-Instruct}, \textit{Meta-Llama-3.1-8B-Instruct} \citep{dubey2024llama}, and four biomedical models derived from previously mentioned general-purpose language models: \textit{BioMistral-7B} \citep{labrak-etal-2024-biomistral}, \textit{Llama3-Med42-8B} \citep{christophe2024med42}, \textit{Llama3-Aloe-8B-Alpha} \citep{gururajan2024aloe}, and \textit{Meditron3-8B} \citep{meditron3}.
We evaluated model performance on the \textbf{MeDiSumQA} dataset through automatic and manual assessments to ensure a comprehensive analysis.

\subsubsection{Automatic Evaluation}

\begin{table*}[htp]\centering
\begin{tabular}{lllllllll}\toprule
Model &Biomedical &Avg &R-L &R-1 &R-2 &BERT F1 &UMLS F1 \\\midrule
Lower Bound &- &20.93 &13.11 &15.76 &2.82 &60.22 &12.74 \\
Upper Bound &- &44.72 &41.55 &45.13 &16.82 &81.35 &38.75 \\\midrule
BioMistral-7B &Yes &23.69 &15.1 &19.67 &5.29 &64.24 &14.13 \\
Llama3-Med42-8B &Yes &29.27 &21.2 &26.84 &8.65 &68.45 &21.21 \\
Llama3-Aloe-8B-Alpha &Yes &19.47 &8.94 &12.11 &3.81 &61.83 &10.66 \\
Meditron3-8B & Yes &29.00 &21.1 &26.63 &8.63 &68.01 &20.62 \\\midrule
Mistral-7B-Instruct-v0.1 &No &23.24 &14.55 &19.00 &5.08 &64.15 &13.42 \\
Meta-Llama-3-8B-Instruct &No &28.75 &20.78 &26.51 &8.72 &67.69 &20.06 \\
Meta-Llama-3.1-8B-Instruct &No &\textbf{31.43} &\textbf{24.1} &\textbf{29.93} &\textbf{10.24} &\textbf{69.35} &\textbf{23.55} \\
\bottomrule
\end{tabular}
\caption{Automatic evaluation of seven models on \textbf{MeDiSumQA}.}\label{tab:automatic-evaluation}
\end{table*}

We evaluated the models using established similarity metrics that capture both n-gram overlap and semantic similarity. The temperature was set to $1.0$ for all models. Due to the long input length, the models were prompted with a one-shot example.  Additional details about the prompts are described in Appendix \ref{sec:appendix:prompts}. 

Specifically, we used ROUGE-1, ROUGE-2, and ROUGE-L \citep{lin2004rouge} to measure lexical overlap at varying levels of granularity, as well as BERT Score \citep{Zhang*2020BERTScore:} to evaluate semantic similarity using contextual embeddings. For the BERT Score we tuned the rescaling baselines for MIMIC-IV discharge summaries using \textit{Bio\_ClinicalBERT} \citep{alsentzer-etal-2019-publicly}. We also used the Unified Medical Language System (UMLS) parser \textit{scispaCy} \citep{neumann-etal-2019-scispacy} to assess the alignment of biomedical entities between predictions and ground truth answers, computing a UMLS F1 score.

As reference for these metrics, we calculated both lower and upper bounds. The lower bound was set by using the input question as the model's prediction, providing a baseline similarity measure. For the upper bound, we reformulated ground truth answers using \textit{Llama-3.3-70B-Instruct} and measured their similarity to the original ground truth.

\subsubsection{Manual Evaluation}

To complement the automatic evaluation, we manually assessed 100 generated answers from two models: \textit{Mistral-7B-Instruct-v0.1}, a lower-scoring model, and \textit{Meta-Llama-3.1-8B-Instruct}, a higher-scoring model. For each model, we sorted the answers by the average similarity score across all automatic metrics. We then divided them into five equal-sized bins, with the lowest $20\%$ placed in bin 1, the next $20\%$ in bin 2, up to bin 5 containing the highest $20\%$. We then sampled ten predictions from each bin.

The answers were rated by a physician on five critical aspects:
\begin{itemize}
    \item \textbf{Factuality:} Accuracy of medical information, rated on a scale from 1 to 5.
    \item \textbf{Brevity:} Conciseness of the response, rated on a scale from 1 to 5.
    \item \textbf{Patient-Friendliness:} Clarity and accessibility of the response for laypersons, rated on a scale from 1 to 5.
    \item \textbf{Relevance:} Alignment of the response with the question, rated on a scale from 1 to 5.
    \item \textbf{Safety:} Potential for harm or dissemination of misleading information, rated as a binary score (unsafe [0]/safe [1]).
\end{itemize}

Using the same sampling scheme and models, we collected 100 additional model-generated answers. These answers were then compared to their ground truth by a physician in a blinded fashion, indicating the preferred answer for each pair.

\section{Results}
\begin{figure*}[h]
  \centering
  \includegraphics[width=\textwidth]{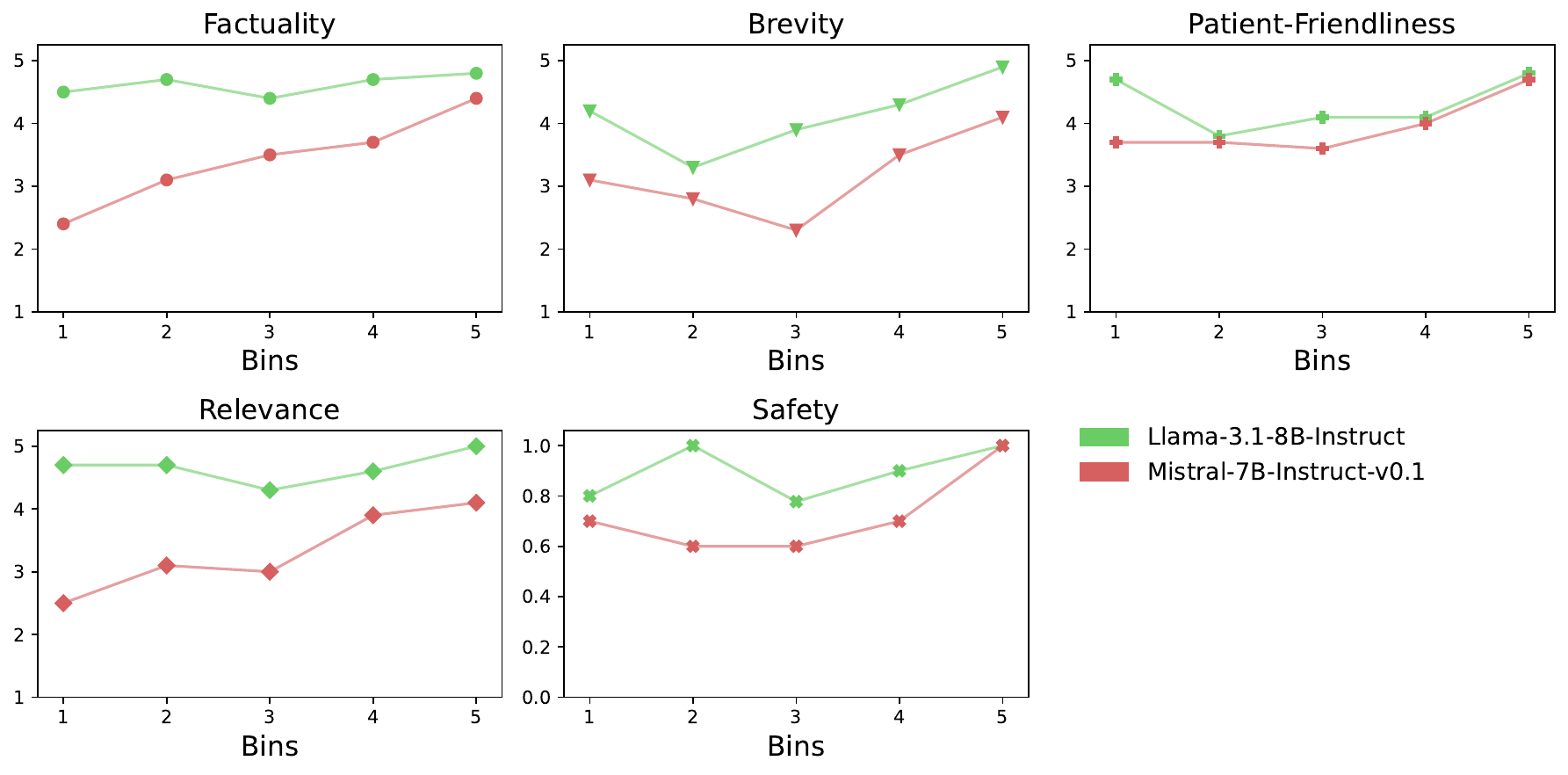}
  \caption{Physicians’ evaluation of model generated answers on \textbf{MeDiSumQA}. Generated answers by \textit{Llama-3.1-8B-Instruct} (green) and \textit{Mistral-7B-Instruct-v0.1} (red) were sorted by their average automatic evaluation scores and divided into 5 bins. From each bin, 10 examples per model were sampled and rated by a physician across \textit{Factuality}, \textit{Brevity}, \textit{Patient-Friendliness}, \textit{Relevance}, and \textit{Safety}. Each subplot displays scores either between 1 and 5 [\textit{Factuality}, \textit{Brevity}, \textit{Patient-Friendliness}, \textit{Relevance}] or 0 and 1 [\textit{Safety}].}
  \label{answer_quality}
\end{figure*}

\subsection{MeDiSumQA Description}
Initially, we generated $500$ QA pairs, which were reduced to 416 pairs after manual curation. Figure \ref{qas_examples} shows three examples of the resulting QA pairs.

Analysis of the QA categories in \textbf{MeDiSumQA} show a fairly even distribution across most categories (Figure \ref{medisumqa_categories}). \textit{Treatment \& Hospital Course} make up the largest portion at $22.4\%$. \textit{Procedures \& Tests}, \textit{Medications}, \textit{Symptoms \& Complications}, and \textit{Diagnosis} each range between $17.1\%$ and $20.9\%$. \textit{Post-Discharge Care \& Follow-Up} questions are notably underrepresented at only $4.8\%$.

\subsection{Automatic Evaluation}

Automatic evaluation across different LLMs reveals varying performance on \textbf{MeDiSumQA} (Table \ref{tab:automatic-evaluation}). 

\textit{Meta-Llama-3.1-8B-Instruct} performed best among all tested metrics, achieving the highest scores despite being a general-domain model without specific biomedical adaptation. 

Comparing biomedical-adapted models with their general-domain counterparts reveals mixed results. Some biomedical adaptations showed only marginal improvements over their base models:  \textit{BioMistral-7B} marginally outperformed its base model \textit{Mistral-7B-Instruct-v0.1} with a small increase of 0.45 points, while \textit{Llama3-Med42-8B} showed a similar pattern with a slight improvement of 0.52 points over \textit{Meta-Llama-3-8B-Instruct}.

However, several biomedical adaptations performed notably worse. Most striking is the case of \textit{Llama3-Aloe-8B-Alpha}, which showed a substantial decrease of 9.28 points compared to its base model \textit{Meta-Llama-3-8B-Instruct}. Similarly, \textit{Meditron3-8B} exhibited a considerable decline of 2.43 points relative to \textit{Meta-Llama-3.1-8B-Instruct}.

\subsection{Manual Evaluation}

Manual comparison of \textit{Llama-3.1-8B-Instruct} and \textit{Mistral-7B-Instruct-v0.1} across factuality, brevity, patient-friendliness, relevance, and safety revealed differences between the lower and higher scoring models (Figure \ref{answer_quality}).

In terms of factuality, \textit{Llama-3.1-8B-Instruct} demonstrated consistently high performance, maintaining scores above 4.0 across all bins, with minimal variation. In contrast, \textit{Mistral-7B-Instruct-v0.1} showed a gradual improvement from bin 1 (score ~2.5) to bin 5 (score ~4.3).

In the brevity metric, both models showed improved scores in higher bins. \textit{Llama-3.1-8B-Instruct} maintained generally higher brevity scores throughout, starting at approximately 4.0 in bin 1 and reaching nearly 5.0 in bin 5. \textit{Mistral-7B-Instruct-v0.1} displayed more variable performance, with a notable dip in bin 3 before recovering in bins 4 and 5.

Patient-friendliness scores converged for both models in the higher bins, with both achieving scores near 4.5 in bin 5. \textit{Llama-3.1-8B-Instruct} showed initially higher scores in the lower bins, while \textit{Mistral-7B-Instruct-v0.1} maintained relatively consistent scores around 3.5 before improving in the higher bins.

Regarding relevance, \textit{Llama-3.1-8B-Instruct} consistently outperformed its counterpart, maintaining scores above 4.5 across all bins. \textit{Mistral-7B-Instruct-v0.1} showed a gradual improvement from approximately 2.5 in bin 1 to 4.0 in bin 5. 

Safety scores for both models were relatively high, with \textit{Llama-3.1-8B-Instruct} showing slightly better performance, particularly in bins 2 and 3.

\begin{figure*}
\centering
\begin{subfigure}{.5\textwidth}
  \centering
  \includegraphics[width=0.9\columnwidth]{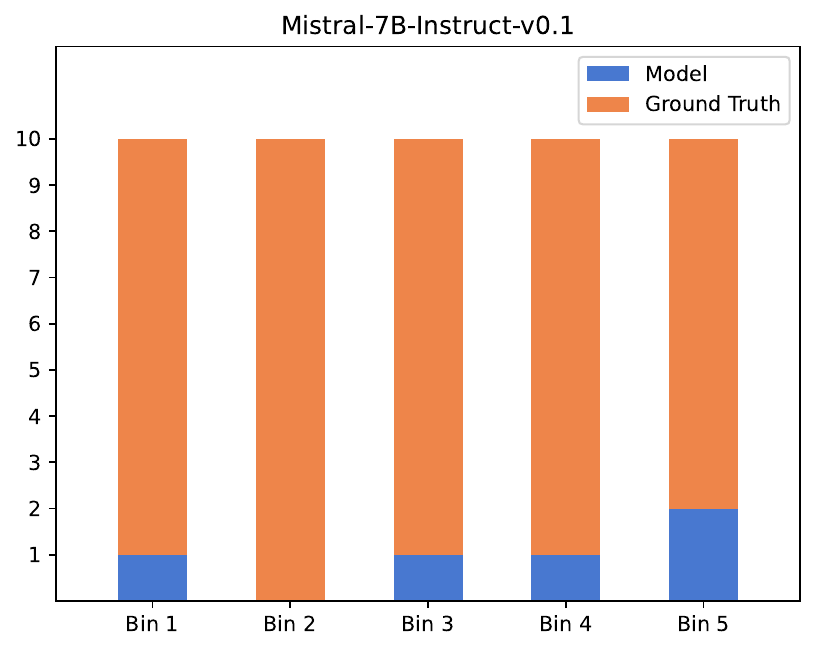}
    \caption{}
  \label{Mistral-7B-Instruct-v0.1_vs_gt}
\end{subfigure}%
\begin{subfigure}{.5\textwidth}
  \centering
  \includegraphics[width=0.9\columnwidth]{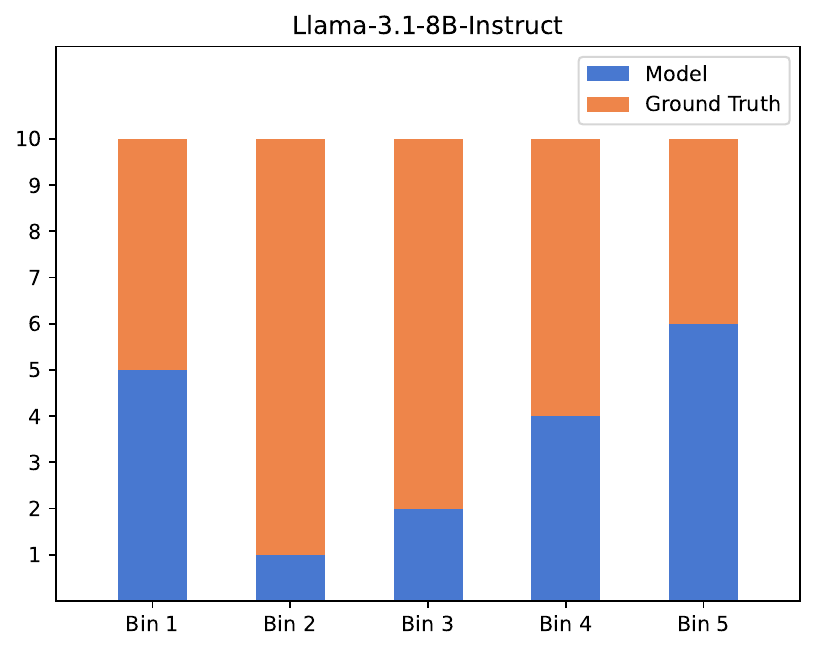}
      \caption{}
  \label{Llama-3.1-8B-Instruct_vs_gt}
\end{subfigure}
\caption{Physician preferences for answers generated by \textit{Mistral-7B-Instruct-v0.1} (a) and \textit{Llama-3.1-8B-Instruct} (b) and the ground truth answers.}
\label{models_vs_gt}
\end{figure*}

When a physician rated preferences between ground truth and model-generated answers, ground truth responses were generally preferred, though the patterns differed between models (Figures \ref{Mistral-7B-Instruct-v0.1_vs_gt}, \ref{Llama-3.1-8B-Instruct_vs_gt}. 

For \textit{Mistral-7B-Instruct-v0.1}, ground truth answers were strongly preferred across all bins, with model-generated answers favored only in exceptional cases.

For \textit{Llama-3.1-8B-Instruct}, the results were more nuanced. Model-generated answers were preferred equally or slightly more often in cases with very high, but also with very low automatic similarity scores. In the middle ranges (bins 2, 3, and 4), ground truth answers were strongly preferred, though model-generated responses still garnered 10–40 \% preference, with higher rates in the upper bins.

\section{Discussion}

Here, we introduce \textbf{MeDiSumQA}, a benchmark dataset designed to evaluate the ability of LLMs to answer clinical questions in a patient-friendly manner. By combining automatic and manual evaluations, our study provides insights into the strengths and limitations of LLMs for patient-oriented question answering, thus narrowing the gap between complex medical information and safe patient communication.

\subsection{Characterization of the Dataset}

\textbf{MeDiSumQA} provides a diverse and structured set of patient-oriented QA pairs derived from discharge summaries, covering key medical topics relevant to patient care. The category distribution of \textbf{MeDiSumQA} indicates comprehensive coverage across six major domains, with a particular emphasis on in-hospital care, medical interventions, and treatment courses. This suggests that the dataset aligns closely with the most immediate concerns patients may have after hospitalization, such as understanding their diagnosis, medications, and follow-up care.

While the dataset captures essential aspects of patient education, \textit{Post-Discharge Care \& Follow-Up} is underrepresented. This imbalance may reflect the structure of discharge summaries themselves, which tend to focus more on inpatient treatment rather than long-term care guidance. Expanding \textbf{MeDiSumQA} to include additional post-discharge documentation, such as outpatient follow-up notes or rehabilitation plans, could improve \textbf{MeDiSumQA}'s ability to support patient education beyond hospital stays.

\subsection{Automatic Evaluation}

\textbf{MeDiSumQA} requires LLMs to perform multiple skills simultaneously. Models must comprehend discharge summaries to understand patient cases, extract relevant details about hospital stays, and present this information in patient-friendly language. The discharge summaries are notably long, averaging 3,245.66 tokens with a standard deviation of 1,419.91, which is a significant challenge for LLMs due to the need for effective long-context reasoning \citep{long-context-llms}. Furthermore, models must possess comprehensive medical knowledge and understanding of clinical guidelines to provide accurate follow-up advice. This complex task therefore evaluates an LLM's ability to integrate comprehension, information extraction, clear communication, and medical expertise in a patient-centered context.

Considering these antecedents, our evaluation shows that general-domain LLMs match or exceed the performance of specialized ones on biomedical tasks. Notably, \textit{Meta-Llama-3.1-8B-Instruct} outperformed all tested biomedical domain-adapted models, raising questions about domain-specific training's effectiveness. While some biomedical models showed slight improvements over their base versions, others experienced significant performance declines, highlighting the inconsistent success of domain adaptation approaches.

These findings suggest that comprehensive pre-training on general-domain data may be more valuable than domain-specific adaptation. This challenges the conventional view that specialized tasks require domain-specific training, aligning with recent research questioning the effectiveness of biomedical adaptation \citep{dada2024doesbiomedicaltraininglead, jeong-etal-2024-medical, dorfner2024biomedical}.

\subsection{Correlation of automatic and manual Evaluation}

When comparing automatic with manual evaluation, our results show that calculated metrics like ROUGE and BERT Score correlate well with human judgment. Higher automated metric scores consistently corresponded to higher manual ratings and preferences, particularly for higher-scoring predictions. Conversely, answers from lower-performing models were rarely preferred by physicians and were sometimes deemed unsafe. This correlation between manual scores and physicians' assessments validates that LLMs can be well assessed in their capability to answer medical questions in a patient-friendly manner using \textbf{MeDiSumQA}. 

However, manual assessment also reveals important limitations of automatic metrics, especially when models generated correct but different responses from the ground truth. Notably, in blind preference tests, \textit{Llama-3.1-8B-Instruct} answers were sometimes preferred over ground truth answers, indicating that LLMs can generate valid alternative responses to the ground truth in \textbf{MeDiSumQA} that may be more appealing. Our manual evaluation also shows that LLMs favor safety over conciseness in their responses. These findings underscore the importance of combining human evaluation with automated scoring for thorough assessment in specialized healthcare applications.

\subsection{Data Contamination}

If evaluation datasets overlap with an LLM’s training data, benchmark validity of these datasets is compromised due to data contamination \citep{Li-2024-arxiv, Deng-2023-arxiv}. Such contamination can cause models to memorize rather than generalize, artificially inflating their performance. Although it is possible that some LLMs in our study have encountered parts of the MIMIC-IV dataset, this is unlikely since MIMIC-IV requires authentication for access.

A broader concern for datasets is intentional benchmark manipulation, when models are deliberately trained on evaluation datasets, which compromises dataset reliability. One solution is to generate datasets using private, inaccessible data. To facilitate this, we offer our dataset generation pipeline as open-source, allowing hospitals and other organizations to create confidential benchmarks from their own clinical reports. By releasing our \textbf{MeDiSumQA} code publicly, we enable others to develop independent datasets and conduct robust LLM evaluations using private medical data.

\subsection{Limitations}

Despite its strengths, \textbf{MeDiSumQA} presents challenges. The dataset primarily focuses on English-language discharge summaries, limiting its applicability to multilingual settings. Additionally, while automated metrics such as ROUGE and BERT Score provide valuable insights, our manual assessments reveal that these do not always align with human judgment, particularly in terms of brevity and relevance. Future research should explore more robust evaluation methods that incorporate real-world patient feedback.

\subsection{Outlook}

We make \textbf{MeDiSumQA} available to the public, which offers an opportunity for widespread adoption in the medical AI community, enabling robust evaluations of models based on their ability to generate accurate, patient-friendly responses. This transparency can drive improvements in patient-centered AI by ensuring models are assessed against expert-validated benchmarks.

During manual evaluation, some model-generated answers were preferred over the ground truth. This presents an opportunity to refine the dataset by incorporating high-quality model-generated responses, with physicians selecting the most appropriate answers. As this approach could introduce bias toward LLMs used in the selection process, future versions of \textbf{MeDiSumQA} could involve multiple independent reviewers to ensure broader generalizability.

Lastly, expanding the dataset by applying our pipeline to a larger set of discharge summaries in different languages would enable use cases beyond single-language few-shot evaluation, including fine-tuning models for improved patient-oriented applications. Making the dataset more diverse and scalable will help develop safer, more effective AI-driven healthcare solutions.

\section{Conclusion}

\textbf{MeDiSumQA} represents another step toward enhancing patient understanding of medical documents by providing benchmarks to assess LLMs in answering medical questions in a patient-friendly manner. By evaluating models on both automated and human-centered metrics, our study demonstrates that automatic metrics correlate well with human judgment while also highlighting the potential of general-purpose LLMs in patient education. By making \textbf{MeDiSumQA} accessible on PhysioNet, we aim to foster further research into the applicability of LLMs for patient-oriented question answering and encourage advancements in this field. We hope that \textbf{MeDiSumQA} will serve as a valuable resource for the development of more patient-friendly AI systems, ultimately bridging the gap between complex medical information and safe, effective patient communication.

\bibliography{anthology,custom}

\appendix

\section{Prompts}
\label{sec:appendix:prompts}
\begin{figure*}[ht]
  \centering
  \includegraphics[width=0.8\textwidth]{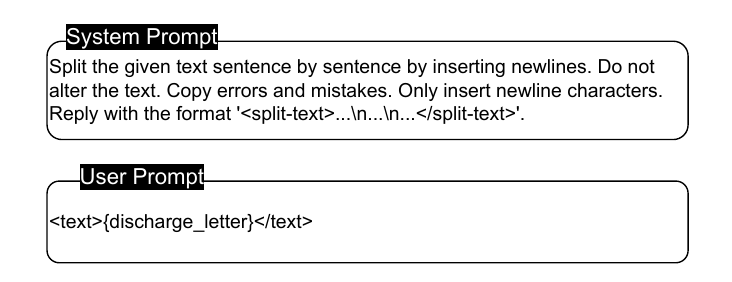}
  \caption{Sentence Splitting prompt}
  \label{sentence_splitting_prompt}
\end{figure*}

\begin{figure*}[ht]
  \centering
  \includegraphics[width=\textwidth]{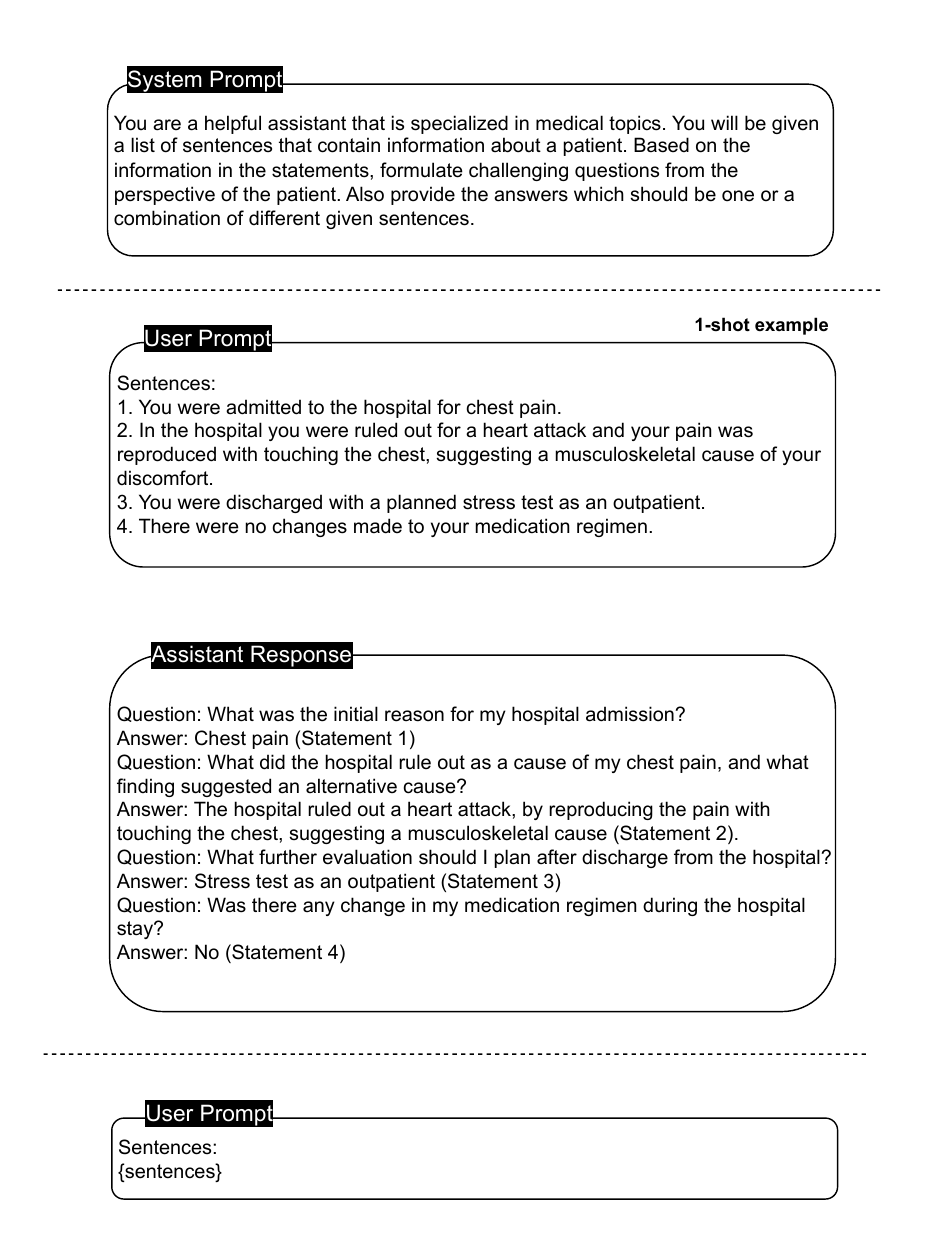}
  \caption{MeDiSumQA generation prompt}
  \label{MeDiSumQA_Generation_Prompts}
\end{figure*}

\begin{figure*}[ht]
  \centering
  \includegraphics[width=\textwidth]{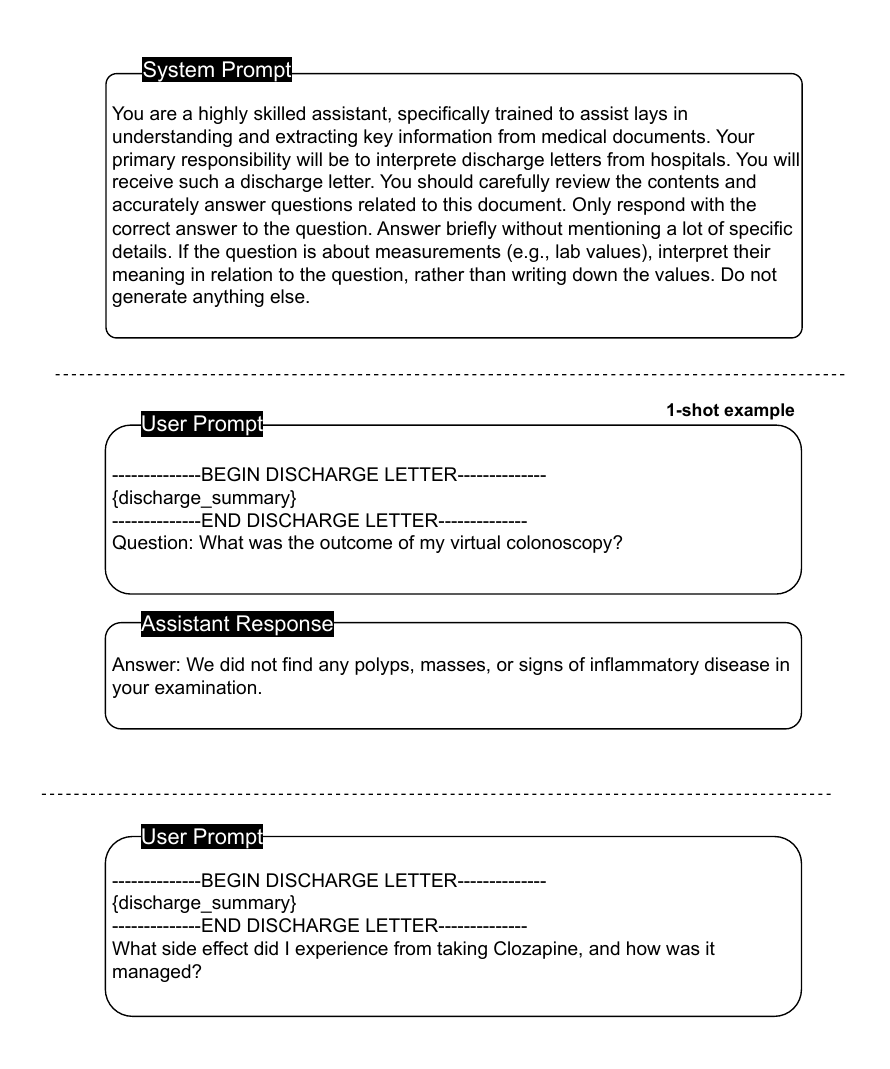}
  \caption{MedisumQA Inference}
  \label{MedisumQA_Inference}
\end{figure*}

Figures \ref{sentence_splitting_prompt} and \ref{MeDiSumQA_Generation_Prompts} show the prompts we use to split the discharge letter into sentences and generate question-answer pairs. For the question-answer generation we include a one shot example. Figure \ref{MedisumQA_Inference} shows the prompt we use to evaluate LLMs on \textbf{MeDiSumQA}.

\end{document}